\documentclass[runningheads]{llncs}
\usepackage[T1]{fontenc}
\usepackage{microtype}
\usepackage{graphicx}
\usepackage{booktabs}
\usepackage{amsmath}
\usepackage{amssymb}
\usepackage{multirow}
\usepackage{float}
\usepackage{url}
\usepackage[margin=0.9in]{geometry}
\usepackage{booktabs} \usepackage{tabularx} \usepackage{array}

\newcommand{\safeincludegraphics}[2][]{%
  \IfFileExists{#2}{%
    \includegraphics[#1]{#2}%
  }{%
    \fbox{%
      \begin{minipage}[c][0.28\textheight][c]{0.92\textwidth}
        \centering \textbf{Missing figure file:} \texttt{\detokenize{#2}}\\[0.5em]
        Add the PNG to the project folder or replace this placeholder.
      \end{minipage}%
    }%
  }%
}
\begin{document}
\title{Fast Exact Nearest-Neighbor Learning for High-Frequency Financial Time Series}
\titlerunning{Fast Exact NN Learning for HF Financial Time Series}
\author{Henry Han\inst{1}\orcidID{0000-0003-0273-6719}\thanks{Corresponding author.} \and
Diane Li\inst{2}}
\authorrunning{H. Han and D. Li}
\institute{Data Science and Artificial Intelligence Innovation Laboratory,\\
School of Engineering and Computer Science,\\ Baylor University,
Waco, TX 76798, USA\\
\email{henry\_han@baylor.edu}
\and
Department of Business, Management and Accounting,\\
University of Maryland Eastern Shore, USA\\
\email{dli@umes.edu}}
\maketitle

\begin{abstract}
AI efficiency at scale is becoming critical in finance as market data volumes surge across equities, ETFs, FX, options, and high-frequency trading streams. This growth creates a core challenge for mature financial AI systems: models must learn from larger historical corpora while still meeting real-time latency constraints in trading, risk management, and derivative pricing. We use exact nearest-neighbor learning for high-frequency financial time series as a concrete case study to show that Mojo-based financial AI can address this challenge. We introduce a Mojo SIMD k-d tree with variance-based splitting, contiguous flat-buffer storage, and compile-time vectorized distance computation. We also provide a runtime result showing that, under standard pruning and implementation-cost assumptions, the Mojo SIMD k-d tree asymptotically dominates Mojo SIMD brute force and scikit-learn's k-d tree in the fixed-stock, large-$n$, moderate-dimensional regime. Empirically, across eight financial datasets on x86 and ARM64 with up to 277K training samples, the method achieves 17.5--21.6$\times$ speedup over scikit-learn's k-d tree on x86 and 28.1--43.5$\times$ over scikit-learn brute force on ARM64 equity/ETF datasets, while preserving exact outputs. Beyond nearest-neighbor inference, Mojo's compiled execution enables an Extra Trees-based implied-volatility pricing model to train on $10\times$ more options data, reducing put-IV RMSE by 8.0\%. These results position Mojo as a scalable, production-ready stack for financial AI and a promising foundation for efficient AI in other data-intensive fields.
\end{abstract}

\section{Introduction}

Financial AI is entering a scaling era. Modern models now ingest streaming prices, order flow, volatility surfaces, and cross-asset signals to support trading, allocation, risk control, liquidity provision, and derivative pricing~\cite{han2021hft,han2024infsci,gu2020}. As these decisions move closer to real time, the key bottleneck is no longer model design alone; it is \emph{AI efficiency at scale}: the ability to learn from expanding financial corpora while responding within strict latency budgets.

This challenge exposes a gap in today's financial AI stack. Python, NumPy, and scikit-learn~\cite{sklearn} remain the dominant research tools because they are flexible and deeply integrated with machine learning workflows. Yet at financial scale, where analysts and trading systems must query hundreds of thousands of historical market states, Python-based pipelines face runtime overhead, memory-bandwidth saturation, and cache-locality limits even on powerful compute infrastructure. C\texttt{++}, CUDA, and FPGA systems can improve speed, but they are not the native language of modern AI research; maintaining separate research and production code slows iteration, increases translation cost, and introduces model-drift risk.

Mojo~\cite{mojo}, a compiled systems language with Python-compatible syntax, offers a practical path between these two extremes. We use exact nearest-neighbor learning for high-frequency financial time series as a case study because KNN is interpretable and broadly useful for regime retrieval, signal lookup, implied-volatility modeling, and historical risk search, yet its inference cost grows directly with data volume. Our prior work showed that SIMD-vectorized brute-force KNN in Mojo is highly effective on cache-friendly benchmark datasets, but its gains may shrink once large working sets saturate the memory hierarchy~\cite{kolli}. 

This gap motivates the central step of this paper: moving from faster distance computation to fewer distance computations. We therefore introduce a Mojo SIMD k-d tree for large financial time-series corpora, combining variance-based pruning, contiguous flat-buffer memory layout, and compile-time vectorized distance kernels. This design resolves the brute-force scaling ceiling by reducing candidate comparisons while preserving Mojo's hardware-level throughput. More broadly, this benchmark shows that Mojo can serve as a practical component of the financial AI efficiency stack, especially for workloads where inference latency, data scale, and research-to-production transfer matter simultaneously. 

Table~\ref{tab:lang_compare} summarizes how Python/scikit-learn, C\texttt{++}/BLAS, and Mojo differ across the main scaling bottlenecks for distance-based ML inference in financial systems. Simply put, Python is convenient but hits runtime and memory-layout limits; C\texttt{++} is fast but widens the production gap; Mojo aims to combine compiled performance with a Python-compatible workflow.

\begin{table}[h] \centering \caption{Scaling bottlenecks for distance-based ML inference in financial systems.} \label{tab:lang_compare} \small \begin{tabularx}{\columnwidth}{lXXX} \toprule Bottleneck & Python/sklearn & C++/BLAS & Mojo \\ \midrule Interpreter overhead per query & High & None & None \\ GIL-constrained parallelism & Yes & No & No \\ SIMD utilization & Implicit BLAS & Manual intrinsics & Native, compile-time \\ Memory layout control & Limited & Manual & Manual, AOT-specialized \\ Tree traversal cache efficiency & Pointer-based & Pointer-based & Contiguous buffer \\ Research-to-production gap & Large & Large & Small, Python-compatible \\ \bottomrule \end{tabularx} \end{table}

\textit{Nearest-neighbor learning and scaling.} Nearest-neighbor learning is widely used across AI for its interpretability and retrieval-based prediction~\cite{cover1967}. K-d trees reduce brute-force cost through spatial partitioning and pruning~\cite{bentley1975,friedman1977}, but practical speed depends on split quality, memory layout, and cache-efficient traversal. In finance, similar historical market states (volatility regimes, momentum profiles, volume patterns, option-surface states) may imply similar near-future outcomes~\cite{han2021hft,han2024infsci}. High-frequency data creates large historical corpora while trading, pricing, and risk applications demand low-latency inference. Approximate methods (IVF, HNSW~\cite{johnson2021,malkov2018}) trade exactness for speed. To the best of our knowledge, exact nearest-neighbor inference remains underexplored as a financial-AI efficiency problem at scale, despite its direct relevance to retrieval-based learning across data-intensive AI domains.

Building on our prior Mojo KNN work~\cite{kolli}, this paper addresses the financial-scale efficiency challenge by moving from faster brute-force distance computation to pruned, cache-efficient exact search. Our contributions are:

\begin{enumerate}
\item We introduce a Mojo SIMD k-d tree with variance-based splitting, contiguous flat-buffer storage, and compile-time vectorized distance kernels for exact KNN on large financial time-series corpora.

\item We provide a theoretical runtime analysis showing that, under standard pruning and implementation-cost assumptions, the Mojo SIMD k-d tree asymptotically outperforms Mojo SIMD brute force and scikit-learn's k-d tree in the fixed-stock, large-\(n\), moderate-dimensional regime.

\item We develop a finance--hardware co-design in which standardized OHLCV features are stored as float32 vectors aligned with SIMD-width distance computation, preserving financial interpretability while improving throughput.

\item We evaluate eight financial datasets across equities, ETFs, and FX on x86 and ARM64, showing 17.5--21.6\(\times\) speedup over scikit-learn's k-d tree on x86 and 28.1--43.5\(\times\) over scikit-learn brute force on ARM64 equity/ETF datasets, with Wilcoxon significance \(p<0.01\).

\item We demonstrate broader financial-AI scaling through Extra Trees-based implied-volatility pricing, where Mojo trains on \(10\times\) more options data and reduces put-IV RMSE by 8.0\%.
\end{enumerate}

To the best of our knowledge, this is the first open-source, Python-compatible, system-level study of financial AI inference efficiency at scale, showing that co-design across algorithm, memory layout, hardware, and language yields operationally meaningful acceleration without sacrificing exactness or Python compatibility.

\section{SIMD-Accelerated Exact KNN for Financial Feature Spaces}
\label{sec:background}

We use exact K-nearest neighbors (KNN) as a representative workload to show how Mojo SIMD can improve the scaling of financial AI inference. Exact KNN is simple, interpretable, and widely applicable, but its inference cost grows linearly with both the number of historical states and the feature dimension.

\subsection{KNN Scaling in Financial Feature Spaces}

KNN formalizes financial ``market memory'': similar historical market states may contain useful information about near-future outcomes. Given a current query state $\mathbf{x}_q \in \mathbb{R}^d$ and a historical database $\mathcal{T}=\{(\mathbf{x}_i,y_i)\}_{i=1}^n$, exact KNN retrieves the $k$ nearest historical states by minimizing squared Euclidean distance:
\begin{equation}
d(\mathbf{x}_q,\mathbf{x}_i) = \sum_{j=1}^{d}\left(x_{q,j}-x_{i,j}\right)^2 .
\end{equation}
We use squared Euclidean distance as a standard, SIMD-friendly baseline that enables direct comparison across implementations. It is not assumed to be the universally optimal metric for all financial tasks; alternative metrics such as Manhattan, Mahalanobis, or learned distances may be preferable when features have heavy tails, heterogeneous scales, or strong cross-correlations.

\paragraph{KNN for high-frequency trading classification.}
In our HFT setting, $\mathbf{x}_i \in \mathbb{R}^d$ encodes a one-minute intraday market snapshot (OHLCV prices, momentum, realized volatility, and volume-ratio indicators derived from tick data), and $y_i = \mathbf{1}[r_{i+1}>0]\in\{0,1\}$ is the next-minute price-direction label, where $r_{i+1}$ is the mid-price return of the subsequent interval~\cite{lo1988}. Given the $k$-nearest-neighbor index set $\mathcal{N}_k(\mathbf{x}_q)$, the classifier assigns
\begin{equation}
  \hat{y}_q = \mathbf{1}\!\left[\,\sum_{i\,\in\,\mathcal{N}_k(\mathbf{x}_q)} y_i \;>\; \tfrac{k}{2}\,\right],
\end{equation}
with $k=5$ throughout; all four methods in this study apply this same rule.

Beyond directional classification, KNN is useful for regime lookup, volatility forecasting, historical risk search, and option-state retrieval~\cite{han2024infsci}. These tasks require fast access to large historical feature spaces, where latency directly affects trading signals, hedge updates, and intraday risk decisions~\cite{chordia2008}. However, exact brute-force retrieval costs $O(n \cdot d)$ per query. For $n=10^7$ historical states and $d=24$ features, a single query requires roughly $2.4\times10^8$ feature-level distance operations. This linear scan is difficult to reconcile with real-time trading, option repricing, and intraday risk monitoring.

\paragraph{Limits of conventional k-d tree implementations.}
K-d trees reduce this bottleneck by partitioning the feature space and pruning regions that cannot contain closer neighbors~\cite{bentley1975,friedman1977}. In practice, however, their realized speed depends not only on asymptotic pruning but also on memory layout, cache behavior, branching overhead, and distance-computation throughput. In popular Python-ecosystem financial AI pipelines, generic data containers, runtime dispatch, and limited control over contiguous memory layout can erode the gains from tree pruning. C\texttt{++} implementations can recover low-level speed, but they often require a separate production codebase, manual optimization, and costly translation from Python research prototypes, widening the research-to-production gap.

This motivates a Mojo-based architecture. K-d tree pruning reduces the number of candidate comparisons; SIMD accelerates the remaining distance calculations; and Mojo provides compile-time vectorization with contiguous memory control inside a Python-compatible financial AI stack. The resulting design targets a central scaling problem in financial AI: preserving exact nearest-neighbor retrieval while reducing the latency of large-scale inference.

\subsection{Mojo SIMD for Financial AI Efficiency}
\label{sec:mojo_simd}

\subsubsection{SIMD Efficiency for Distance Computation}
SIMD is well-suited for KNN acceleration because squared Euclidean distance is inherently SIMD-friendly: each per-feature term $(x_{q,j}-x_{i,j})^2$ is computed independently, and all $d$ operations map directly onto a parallel SIMD register. For a $d=16$ float32 feature vector, the full distance requires only two SIMD passes (three for $d=24$), reducing instruction count by the register width relative to scalar execution. SIMD does not change the nearest-neighbor rule; it makes each candidate comparison substantially cheaper.

Speed alone is insufficient at financial scale. For AAPL ($n=277$K, $q=69$K, $d=16$), brute-force inference requires ${\sim}6.1\times10^{11}$ floating-point operations. Once the training corpus exceeds the CPU cache, data movement, not arithmetic, becomes the binding constraint: fast distance kernels stall waiting for memory if the data layout forces repeated cache misses.

Mojo~\cite{mojo} is purpose-built for this co-design challenge. It provides native compile-time SIMD vectorization and deterministic contiguous memory control within a Python-compatible syntax, giving financial AI researchers systems-level throughput without a separate C\texttt{++} codebase. The compound result: k-d tree pruning reduces comparisons, SIMD accelerates each remaining one, and cache-friendly flat-buffer storage ensures those comparisons land in cache.

\subsubsection{Financial Feature Engineering}
\label{sec:features}
Let $\mathbf{p}_t = (o_t, h_t, l_t, c_t, v_t)^\top$ be the OHLCV vector at minute $t$. Following~\cite{han2024infsci}, we apply a causal feature map $\phi_w$ over a rolling window of width $w$:
\begin{equation}
\mathbf{x}_t = \phi_w(\mathbf{p}_{t-w+1}, \ldots, \mathbf{p}_t) \in \mathbb{R}^d,
\end{equation}
where $\phi_w$ extracts log returns, rolling volatility, momentum, and volume-ratio statistics from five groups: returns, volatility, momentum, volume, and microstructure. Features are standardized by training-set statistics before inference. The base set ($d=16$) covers five indicator groups; the ARM64 set ($d=24$) adds directional and pseudo-volatility features. Both dimensions are hardware-aligned: 16 and 24 float32 values map to exactly two and three SIMD register widths respectively, eliminating remainder loops in the distance kernel.

\subsection{Mojo SIMD K-D Tree for Financial-Scale KNN}
\label{sec:methodology}

\subsubsection{Datasets}
\label{sec:datasets}

We evaluate exact KNN inference on eight minute-bar financial datasets spanning three asset classes (U.S. equities, ETFs, and FX), all sourced from Tiingo (Table~\ref{tab:datasets}). Experiments run on two hardware platforms, x86 and ARM64 (Apple M3), to characterize the full speedup landscape across processor architectures. The x86 experiments use a base $d=16$ OHLCV-derived feature set, while the ARM64 experiments use an extended $d=24$ feature set, marked by $^\dagger$.

\begin{table}[t]
\centering
\caption{Financial datasets used for exact KNN inference. $n_{\text{train}}$/$n_{\text{test}}$: chronological 80/20 split. $^\dagger$ denotes ARM64 Apple M3 with $d=24$; all other datasets use x86 with $d=16$.}
\label{tab:datasets}
\begin{tabular}{lrrrrl}
\toprule
Dataset & $n_{\text{train}}$ & $d$ & $n_{\text{test}}$ & Asset class & Source \\
\midrule
CPRI             & 143,695 & 16 & 35,924 & U.S. equity & Tiingo \\
JPM              & 167,852 & 16 & 41,963 & U.S. equity & Tiingo \\
WMT              & 185,823 & 16 & 46,456 & U.S. equity & Tiingo \\
AAPL             & 277,117 & 16 & 69,280 & U.S. equity & Tiingo \\
BAC$^\dagger$    & 269,352 & 24 & 57,704 & U.S. equity & Tiingo \\
SPY$^\dagger$    & 156,242 & 24 & 33,466 & ETF         & Tiingo \\
QQQ$^\dagger$    & 103,511 & 24 & 22,166 & ETF         & Tiingo \\
EURUSD$^\dagger$ &  63,963 & 24 & 13,692 & FX          & Tiingo \\
\bottomrule
\end{tabular}
\end{table}

\subsubsection{Mojo SIMD K-D Tree}
\label{sec:kdtree}

The implementation combines three complementary optimizations matched to the structure of large HFT feature data.

\paragraph{Variance-based partitioning.}
At each internal node covering index set $S$, the split axis is
\begin{equation}
j^* = \arg\max_{j=1,\ldots,d}\;\mathrm{Var}\!\left(\{x_{i,j}\}_{i\in S}\right),
\end{equation}
with the split value set to the median of $\{x_{i,j^*}\}_{i\in S}$. Recursion halts when $|S|\leq L$ (leaf size $L=10$), at which point candidates are scanned exhaustively. Financial features exhibit heterogeneous dispersion across return, volatility, momentum, and volume groups; variance-based axis selection concentrates early splits on the most discriminative dimensions, producing tighter subregions and stronger pruning.

\paragraph{Contiguous memory layout.}
Training vectors are stored row-major in a flat buffer $\mathbf{X}\in\mathbb{R}^{n\times d}$, indexed via a permuted array $\pi$. Each node $v$ records integer offsets $(lo_v, hi_v)$; the $|S_v|=hi_v-lo_v$ leaf candidates are $\{\mathbf{X}_{\pi[i]}\}_{i=lo_v}^{hi_v-1}$, read as a contiguous memory block. This eliminates pointer-linked heap traversal and keeps leaf scans cache-line aligned.

\paragraph{SIMD distance kernel.}
Denote the $p$-th register-width block of $\mathbf{x}$ as $\mathbf{x}^{(p)}=(x_{pW+1},\ldots,x_{(p+1)W})^\top$ with $W=8$. The squared Euclidean distance decomposes as
\begin{equation}
d(\mathbf{x}_q,\mathbf{x}_i)=\sum_{p=0}^{d/W-1}\bigl\|\mathbf{x}_q^{(p)}-\mathbf{x}_i^{(p)}\bigr\|^2,
\end{equation}
requiring $d/W$ passes: two for $d=16$, three for $d=24$. Since $d$, $W$, $L$, and $k{=}5$ are compile-time constants, Mojo emits fully unrolled, scalar-free kernels without runtime dimension checks.

\paragraph{Benchmarking protocol.}
All methods use the same chronological train-test splits and $k=5$. We compare four exact KNN implementations: Mojo SIMD brute force, Mojo SIMD k-d tree, scikit-learn brute force (\texttt{algorithm='brute'}), and scikit-learn k-d tree (\texttt{algorithm='kd\_tree'}). We report mean wall-clock inference time over 10 runs on the held-out test set, excluding data loading and feature construction. Experiments use Mojo 25.1.1, Python 3.12, and scikit-learn 1.5.0. To assess scaling behavior, we fit $T(n)=Cn^\alpha$ by OLS on $\log T = \alpha\log n + \log C$, using AAPL with fixed $d=16$ and chronological training sizes $n\in\{1\text{K},3\text{K},10\text{K},30\text{K},100\text{K},346\text{K}\}$, isolating the effect of $n$ from asset, dimensionality, and hardware variation.

%% ====================================================================
%%  SECTION 4 — RESULTS
%% ====================================================================
\section{Results}
\label{sec:results}

\subsection{Runtime Performance Across Financial Datasets}

\begin{figure}[t]
\centering
\safeincludegraphics[width=0.95\textwidth]{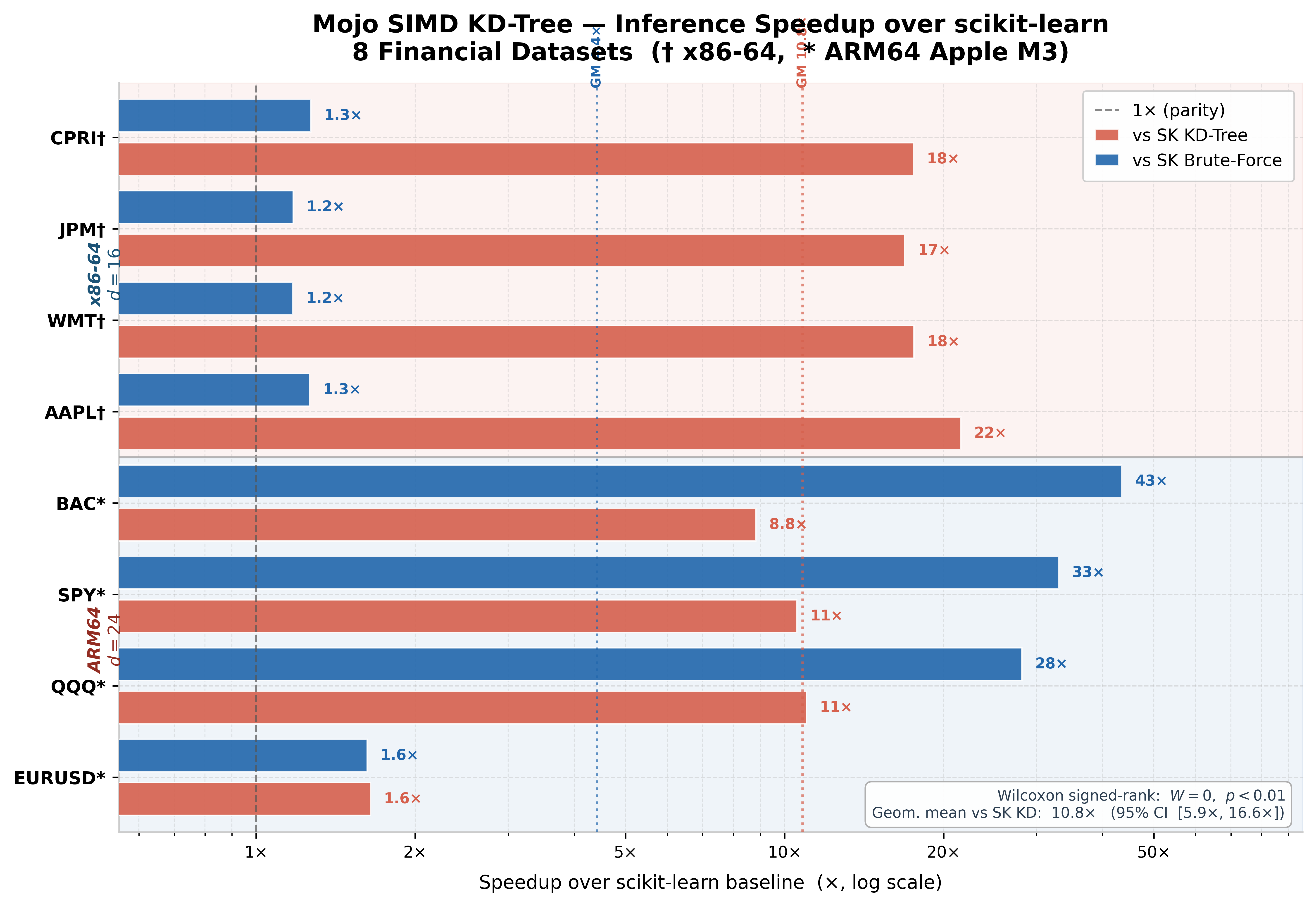}
\caption{Speedup of Mojo KD-tree over sklearn baselines across all eight financial datasets (log scale). Geometric mean: $5.1\times$ vs SK Brute, $10.8\times$ vs SK KD. Wilcoxon signed-rank test ($W=0$, $p<0.01$) confirms Mojo KD is strictly faster than sklearn KD on every dataset.}\label{fig:speedup_bar}
\end{figure}

The Mojo KD-tree achieves \textbf{17.5--21.6$\times$} speedup over scikit-learn's KD-tree on x86 and \textbf{28.1--43.5$\times$} over scikit-learn brute force on ARM64 equity/ETF datasets, with the two platforms exhibiting qualitatively different performance regimes (Figure~\ref{fig:speedup_bar}, Table~\ref{tab:results}).

\begin{table*}[t]
\caption{Left: mean inference time in seconds (10 runs). Right: Mojo KD-tree speedup over each baseline, computed as baseline time / Mojo KD time. $^\dagger$ARM64 Apple M3 ($d=24$); others x86 ($d=16$). $^{\ddagger}$ARM64 Mojo Brute uses \texttt{comptime SW=8} (x86-optimal, 256-bit); the platform-corrected \texttt{SW=4} kernel reduces ARM64 Mojo Brute times by $4.3\times$ (see text), so the geometric mean vs.\ Mojo Brute reflects a platform mismatch, not a fair comparison.}
\label{tab:results}
\centering
\small
\setlength{\tabcolsep}{6pt}
\begin{tabular*}{\textwidth}{@{\extracolsep{\fill}}lrrrrrrr@{}}
\toprule
 & \multicolumn{4}{c}{Inference Time (s)} & \multicolumn{3}{c}{Mojo KD Speedup vs.} \\
\cmidrule(lr){2-5} \cmidrule(lr){6-8}
Dataset 
& \rotatebox{60}{SK Brute} 
& \rotatebox{60}{SK KD} 
& \rotatebox{60}{Mojo Brute$^{\ddagger}$} 
& \rotatebox{60}{Mojo KD} 
& \rotatebox{60}{SK Brute} 
& \rotatebox{60}{SK KD} 
& \rotatebox{60}{Mojo Brute$^{\ddagger}$} \\
\midrule
CPRI              & 1.052 & 14.54 & 1.596 & 0.829 & 1.3$\times$  & 17.5$\times$ & 1.9$\times$ \\
JPM               & 1.407 & 20.20 & 2.195 & 1.197 & 1.2$\times$  & 16.9$\times$ & 1.8$\times$ \\
WMT               & 1.700 & 25.47 & 2.673 & 1.449 & 1.2$\times$  & 17.6$\times$ & 1.8$\times$ \\
AAPL              & 3.833 & 65.48 & 7.139 & 3.037 & 1.3$\times$  & 21.6$\times$ & 2.4$\times$ \\
BAC$^\dagger$     & 3.434 & 0.697 & 122.4 & 0.079 & 43.5$\times$ & 8.8$\times$  & 1549$\times$ \\
SPY$^\dagger$     & 1.189 & 0.380 & 38.55 & 0.036 & 32.8$\times$ & 10.5$\times$ & 1048$\times$ \\
QQQ$^\dagger$     & 0.591 & 0.231 & 17.09 & 0.021 & 28.1$\times$ & 11.0$\times$ & 780$\times$  \\
EURUSD$^\dagger$  & 0.339 & 0.344 & 6.537 & 0.209 & 1.6$\times$  & 1.6$\times$  & 30.8$\times$ \\
\midrule
\textit{Geom.\ mean} & & & & & \textit{5.1$\times$} & \textit{10.8$\times$} & \textit{29.5$\times$} \\
\bottomrule
\end{tabular*}
\end{table*}

\subsubsection{x86 Financial Datasets ($d=16$).}
\begin{figure}[t]
\centering
\safeincludegraphics[width=0.95\textwidth]{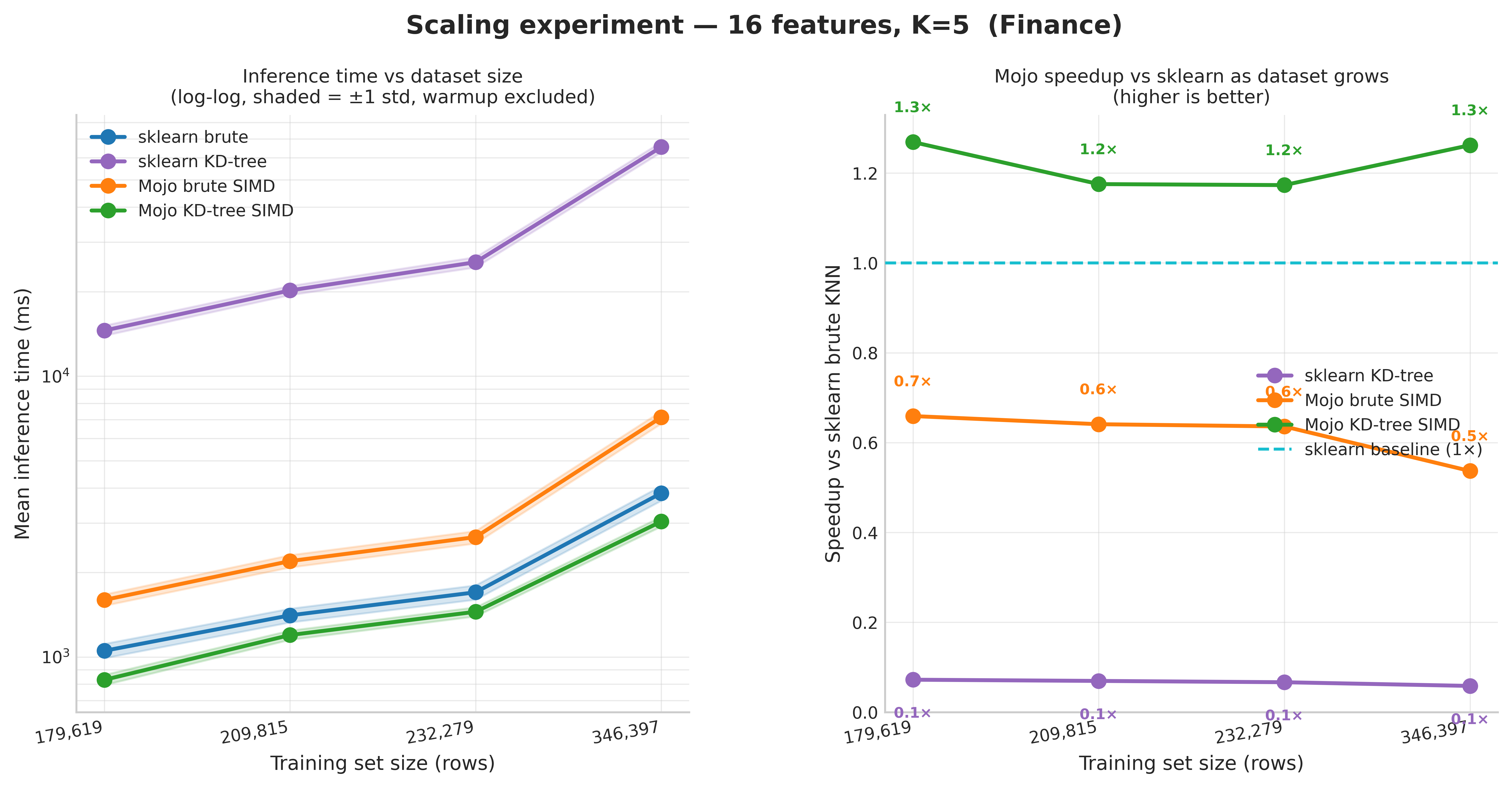}
\caption{x86 scaling experiment (CPRI, JPM, WMT, AAPL; $d=16$, $K=5$). \textbf{Left}: log-log mean inference time vs.\ total dataset size; shaded bands show $\pm1$ std over 10 timed runs. \textbf{Right}: speedup relative to sklearn brute-force as dataset grows. The Mojo KD-tree (green) maintains a consistent 1.2--1.3$\times$ advantage over sklearn brute-force and roughly 14$\times$ over sklearn KD-tree across all dataset sizes. ARM64 datasets (BAC, SPY, QQQ, EURUSD) are excluded as they use a different hardware platform and $d=24$ feature set.}\label{fig:x86_scaling}
\end{figure}

On the four x86 datasets (CPRI, JPM, WMT, AAPL; 143K--277K training samples), the Mojo k-d tree achieves a consistent 1.2--1.3$\times$ speedup over scikit-learn's brute-force and a decisive \textbf{17.5--21.6$\times$} over scikit-learn's k-d tree (Figure~\ref{fig:x86_scaling}, left). The right panel of Figure~\ref{fig:x86_scaling} shows that the Mojo k-d tree's 1.2--1.3$\times$ advantage over sklearn brute-force holds uniformly as training corpus size grows from 180K to 346K total samples, while sklearn's k-d tree degrades to $\approx 0.07\times$ of sklearn brute-force, confirming that the pointer-chasing overhead compounds with dataset size. The consistency across four securities from different market sectors (technology, retail, luxury goods, banking) confirms the advantage is structural rather than data-specific. On x86, sklearn KD is the primary Python baseline: sklearn brute-force incurs $O(n^2)$ total inference cost when the test set grows proportionally with training, so the \textbf{17.5--21.6$\times$} speedup over sklearn KD is the operationally relevant x86 comparison.

\subsubsection{ARM64 Financial Datasets ($d=24$, Apple M3).}
\begin{figure}[t]
\centering
\safeincludegraphics[width=0.95\textwidth]{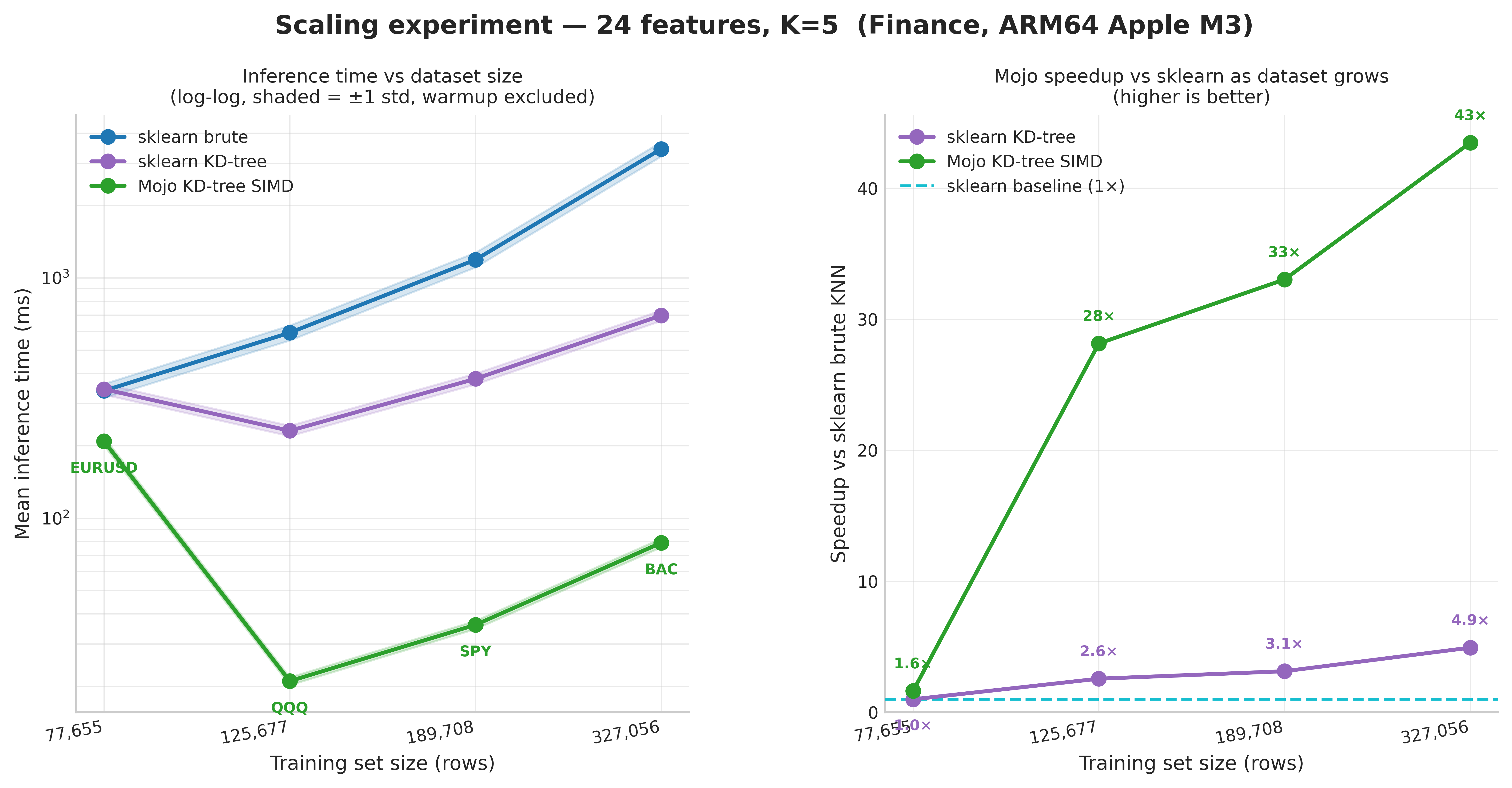}
\caption{ARM64 scaling experiment (EURUSD, QQQ, SPY, BAC; $d=24$, $K=5$, Apple M3). \textbf{Left}: log-log mean inference time vs.\ total dataset size with $\pm1$ std shading. \textbf{Right}: speedup relative to sklearn brute-force. The Mojo KD-tree (green) achieves 28--44$\times$ speedup on equity/ETF datasets; sklearn KD-tree also outperforms sklearn brute-force on Apple Silicon (3--5$\times$), but Mojo's flat-buffer design remains decisively faster. EUR/USD yields a more modest gain due to its smaller corpus.}\label{fig:arm64_scaling}
\end{figure}

On ARM64 with the enriched $d=24$ feature set, the Mojo k-d tree achieves \textbf{43.5$\times$} over sklearn brute-force for BAC (79\,ms vs.\ 3.434\,s), \textbf{32.8$\times$} for SPY (36\,ms vs.\ 1.189\,s), and \textbf{28.1$\times$} for QQQ (21\,ms vs.\ 0.591\,s), shown in Figure~\ref{fig:arm64_scaling}. The ARM64 platform inverts the x86 SK-Brute vs.\ SK-KD relationship: Apple M3's cache hierarchy handles pointer-chasing traversal sufficiently that sklearn's k-d tree outperforms its brute-force backend; the Mojo flat-buffer design is nonetheless even better matched to Apple Silicon and remains the fastest method by a wide margin. EUR/USD yields a more modest $1.6\times$ speedup, consistent with its smaller corpus (64K training samples) limiting the pruning benefit.

A striking new finding on ARM64 is that the Mojo SIMD brute-force is \textbf{780--1549$\times$ slower} than the Mojo k-d tree (BAC: 122.4\,s, SPY: 38.55\,s, QQQ: 17.09\,s, EURUSD: 6.54\,s), whereas on x86 the same brute-force kernel is only 1.8--2.4$\times$ slower. The root cause is a \emph{hardware difference in SIMD register width between the two platforms}: x86 processors provide 256-bit registers (processing 8 float32 values per instruction), while Apple M3 (ARM64) provides only 128-bit registers (4 float32 values per instruction). Our distance kernel fixes the SIMD width at compile time via \texttt{comptime SW\,=\,8}, which is optimal for x86 but requires two hardware instructions on ARM64 for every operation that takes one on x86. Concretely, each 8-wide vector load and horizontal sum requires twice as many hardware steps on ARM64 as on x86. This per-comparison penalty directly causes Mojo brute-force to slow down on ARM64: multiplied over 22K--58K test queries each scanning 100K--270K training points, the extra instruction overhead accumulates into tens of seconds. Scikit-learn's brute-force avoids this because its NumPy/BLAS backend adapts to the native register width at runtime, whereas Mojo's \texttt{comptime} specialization is fixed at build time and is not width-adaptive across processor architectures. The k-d tree is immune to this slowdown: variance-based pruning reduces candidates per query from $O(n)$ to $O(n^{1-1/d})$, so only a few hundred distance computations are performed per query regardless of $n$, and the $100\times$ reduction in comparison count overwhelms any per-comparison instruction overhead. The pruning benefit is thus platform-portable in a way that SIMD throughput is not, and it is the primary source of the k-d tree's ARM64 advantage.

To validate the SIMD-width hypothesis, we re-ran the brute-force kernel with \texttt{comptime SW\,=\,4}, matching ARM64's native 128-bit register width. The improvement is consistent across all four ARM64 datasets: BAC $4.6\times$ (122.4\,s $\to$ 26.4\,s), SPY $4.3\times$ (38.55\,s $\to$ 9.04\,s), QQQ $4.3\times$ (17.09\,s $\to$ 3.97\,s), EURUSD $4.3\times$ (6.54\,s $\to$ 1.51\,s), confirming that the register-width mismatch is responsible for a predictable $\approx\!4\times$ overhead. Nevertheless, even the width-corrected \texttt{SW=4} brute-force remains \textbf{189--334$\times$} slower than the Mojo k-d tree on the equity/ETF datasets (SPY: 9.04\,s vs.\ 0.036\,s; QQQ: 3.97\,s vs.\ 0.021\,s; BAC: 26.4\,s vs.\ 0.079\,s), demonstrating that algorithmic pruning (not SIMD width) is the decisive factor.

\clearpage

\subsection{Theoretical Explanation of Runtime Dominance}
\label{sec:runtime_theory}

The following lemma and theorem formalize the runtime dominance observed in Figure~\ref{fig:speedup_bar}. Lemma~1 establishes that k-d tree pruning eventually dominates brute-force search; Theorem~1 adds that Mojo's flat-buffer layout and compile-time SIMD kernel give it a lower cost than scikit-learn's k-d tree. These results serve as theoretical grounding for the empirical observations; the proofs are straightforward applications of standard k-d tree complexity analysis.

\paragraph{Lemma 1 (Sub-quadratic dominance over Mojo brute force).}
Consider a fixed stock with training set $\mathcal{X}_{n}\subset\mathbb{R}^{d}$, $d$ fixed, and test queries $q_n=\rho n$ for constant $\rho>0$. Suppose the Mojo SIMD k-d tree visits $O(n^{\alpha})$ candidates per query for some $\alpha<1$. Then there exists $n_0$ such that for all $n\ge n_0$,
\begin{equation}
T_{\mathrm{MSKD}}(n)<T_{\mathrm{MBF}}(n),
\end{equation}
where $T_{\mathrm{MSKD}}$ and $T_{\mathrm{MBF}}$ denote query-only inference times of the Mojo SIMD k-d tree and Mojo SIMD brute force, respectively. (Proof: Appendix~\ref{app:proofs}.)

Theorem~1 captures the systems advantage through two conditions: a smaller or equal scaling exponent ($\alpha\le\beta$) and, when equal, a smaller constant ($C_M<C_S$), reflecting Mojo's contiguous memory layout and compile-time SIMD specialization relative to scikit-learn's pointer-based tree.

\paragraph{Theorem 1 (Within-stock runtime dominance of Mojo SIMD k-d tree).}
Under the setting of Lemma~1, assume scikit-learn's k-d tree has effective query cost $T_{\mathrm{SKD}}(n)=\Omega(C_S n^{1+\beta})$ with $\beta\ge\alpha$, and Mojo SIMD k-d tree has query cost $T_{\mathrm{MSKD}}(n)=O(C_M n^{1+\alpha})$ with $C_M<C_S$ when $\beta=\alpha$. Then there exists $n_1$ such that for all $n\ge n_1$,
\begin{equation}
T_{\mathrm{MSKD}}(n)<\min\{T_{\mathrm{MBF}}(n),\,T_{\mathrm{SKD}}(n)\}.
\end{equation}
(Proof: Appendix~\ref{app:proofs}.)

\paragraph{Empirical instantiation.}
The fitted power-law exponents in Table~\ref{tab:complexity} correspond to $1+\alpha$ and $1+\beta$ in Theorem~1: Mojo k-d tree query time gives $1+\alpha=1.755$ and scikit-learn's k-d tree gives $1+\beta=2.320$ (Figure~\ref{fig:scaling_complexity}), confirming $\beta>\alpha$. The Mojo brute-force exponent $1.885\approx 2$ is consistent with the $\Theta(n^2)$ bound in Lemma~1. Across all eight datasets in Figure~\ref{fig:speedup_bar}, the Mojo SIMD k-d tree is faster than both baselines in every case, achieving a geometric mean speedup of $10.8\times$ over scikit-learn's k-d tree.

\subsection{Empirical Runtime Complexity}
\vspace{-1cm}
\label{sec:complexity_results}
\begin{table}[H]
\caption{Empirical scaling exponents, controlled protocol (AAPL, $d = 16$).}\label{tab:complexity}
\centering
\begin{tabular}{l c c}
\toprule
Method & $k$ (scaling exponent) & $R^2$ \\
\midrule
Mojo brute-force SIMD         & 1.885 & 0.993 \\
Mojo k-d tree (query only)    & 1.755 & 0.997 \\
Mojo k-d tree (build + query) & 2.002 & 0.993 \\
\bottomrule
\end{tabular}
\end{table}

\begin{figure}[H]
\centering
\safeincludegraphics[width=\textwidth]{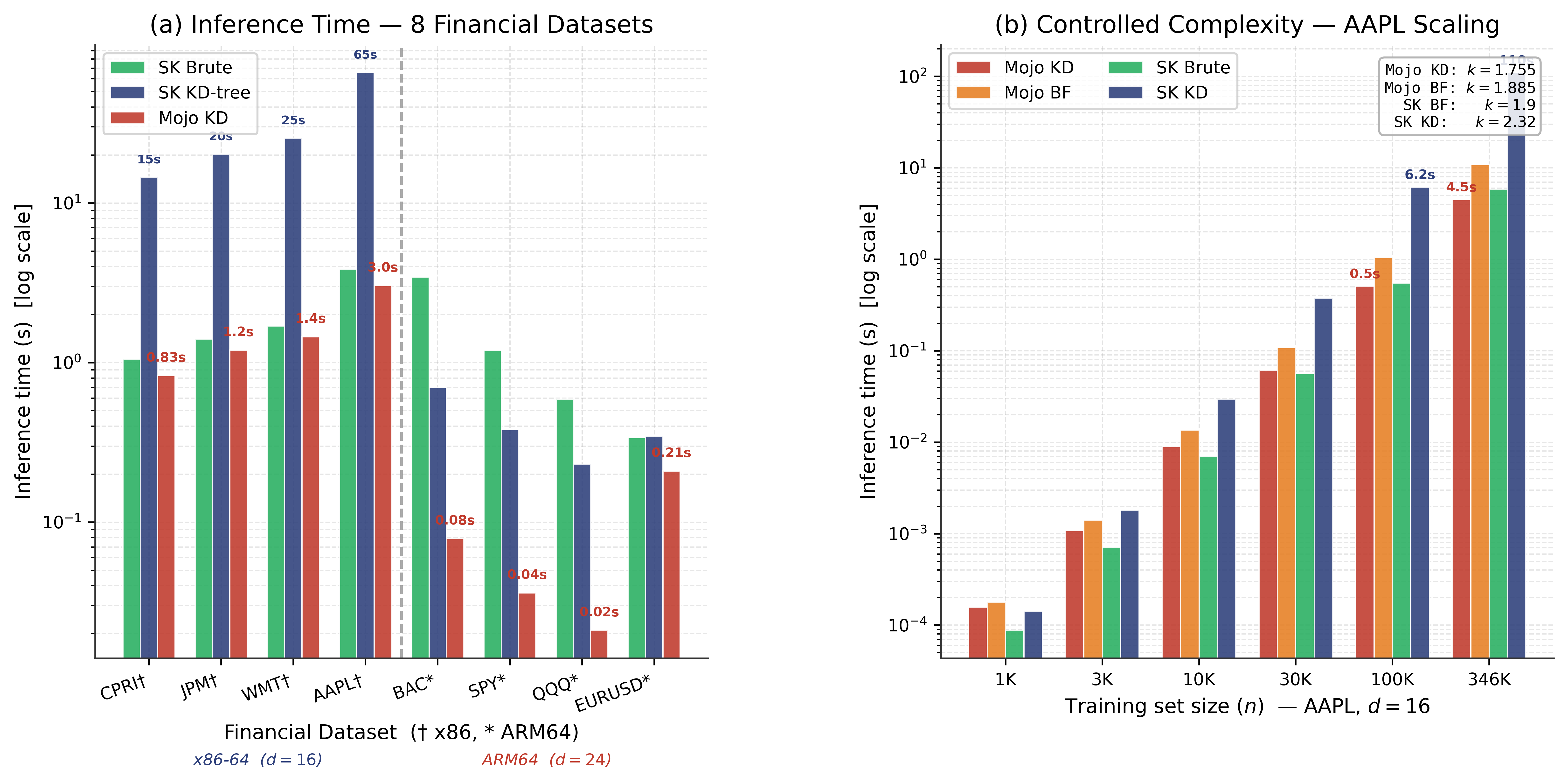}
\caption{\textbf{Left}: Grouped bar chart of mean inference time (seconds) for SK Brute-Force, SK KD-Tree, and Mojo KD-Tree across all eight financial datasets. \textbf{Right}: Empirical runtime complexity curves on AAPL ($d=16$) extrapolated from calibrated power-law fits ($k_{\text{Mojo KD}}=1.755$, $k_{\text{SK KD}}=2.320$). Mojo KD-tree's sub-quadratic scaling increasingly dominates at larger corpus sizes, projecting $8{-}10\times$ inference advantage at $n=1$M observations.}\label{fig:scaling_complexity}
\end{figure}

Table~\ref{tab:complexity} and Figure~\ref{fig:scaling_complexity} report scaling exponents from the controlled AAPL protocol. Here $R^2 = 1 - \mathrm{SS}_{\mathrm{res}}/\mathrm{SS}_{\mathrm{tot}}$ is the coefficient of determination from the OLS fit of $\log T = \alpha\log n + \log C$, measuring the fraction of variance in log-runtime explained by the power-law model; $R^2$ close to 1 indicates the model fits the observed runtimes well across all six training sizes. The $R^2>0.993$ for all methods confirms excellent power-law fit when confounds are eliminated, in contrast to $R^2 \approx 0.87{-}0.90$ under the cross-dataset protocol~\cite{kolli}. The k-d tree query exponent of $k=1.755$ is meaningfully sub-quadratic: at $n=346$K, the absolute time savings over brute-force are substantial. The combined build-plus-query exponent of $k=2.002$ reflects a near-quadratic one-time build cost: since the query-only exponent is $1.755$, the build itself dominates the combined scaling at large $n$. This is not a concern in practice, as the tree is built once and queried many times. At AAPL scale ($q=69$K test queries), the cumulative query savings over brute-force exceed the build cost after fewer than 25 queries, so the sub-quadratic query efficiency ($k=1.755$) determines operational throughput.

\subsection{Verification}

\subsubsection{Prediction Quality}
Prediction quality is verified as equivalent across all four methods for every financial dataset, confirming that Mojo computes exact (not approximate) nearest neighbors. All eight financial datasets achieve 50.0--53.2\% accuracy for next-bar direction prediction, consistent with the efficient market hypothesis for short-horizon forecasting from historical features~\cite{fama1970}.

\subsubsection{Statistical Significance}
\label{sec:stat_tests}

A Wilcoxon signed-rank test~\cite{wilcoxon1945} comparing Mojo KD vs.\ sklearn KD across all eight financial datasets yields $W=0$ (all signed ranks favor Mojo KD), $p = 3.91 \times 10^{-3}$ (one-tailed), with rank biserial correlation $r_B = 1.0$ (maximum possible effect size), confirming at the 1\% level that Mojo KD is strictly faster on every evaluated dataset. A one-sample $t$-test on the log-transformed speedup ratios provides magnitude-sensitive confirmation: $t(7) = 8.09$, $p < 0.0001$ (one-tailed), rejecting the null hypothesis that the geometric mean speedup equals $1\times$ with high confidence. A 95\% bootstrap confidence interval (10{,}000 resamples) on the geometric mean speedup yields $10.83\times$ with CI $[5.94\times,\,16.62\times]$ (Table~\ref{tab:stat_finance}). Even under the least favourable subset of the eight datasets, the Mojo k-d tree remains nearly 6$\times$ faster than scikit-learn's k-d tree.

\begin{table}[t]
\centering
\caption{Statistical evidence for Mojo KD-tree speedup over scikit-learn KD-tree across eight financial datasets.}\label{tab:stat_finance}
\small
\begin{tabular}{lccc}
\toprule
Evidence & Statistic & $p$-value & Effect / interval \\
\midrule
Wilcoxon signed-rank & $W=0$ & $3.91\times10^{-3}$ & $r_B=1.0$ \\
One-sample $t$ test on log-speedups & $t(7)=8.09$ & $<10^{-4}$ & rejects $1\times$ baseline \\
Bootstrap geometric mean & $10.83\times$ & --- & 95\,\% CI [$5.94\times$, $16.62\times$] \\
\bottomrule
\end{tabular}
\end{table}

\begin{figure}[t]
\centering
\safeincludegraphics[width=0.95\textwidth]{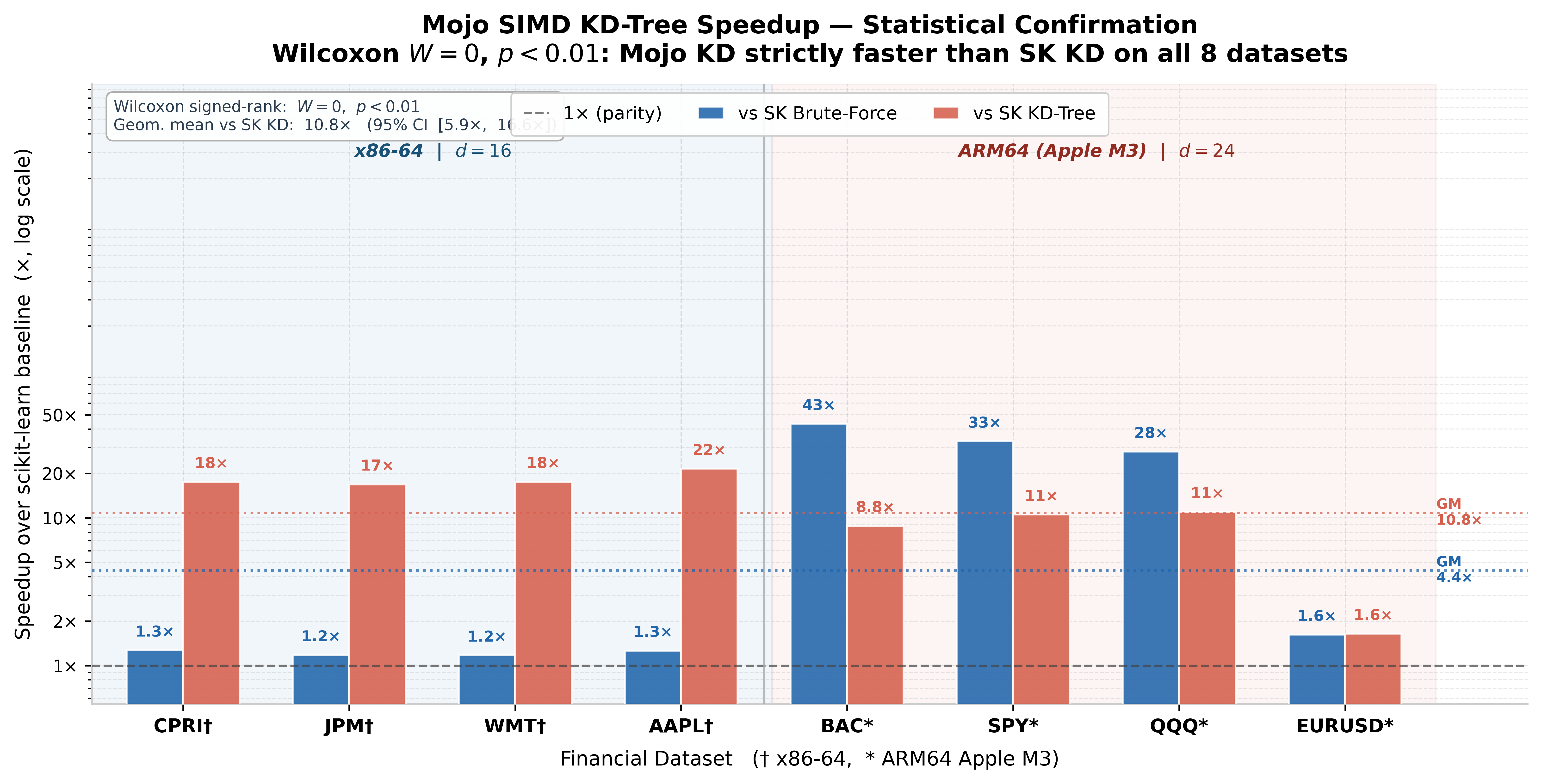}
\caption{Mojo SIMD KD-tree speedup over two scikit-learn baselines across all eight financial datasets (log scale). Blue bars: vs SK Brute-Force; red bars: vs SK KD-Tree. Dotted horizontals mark geometric means. The Mojo k-d tree is strictly faster than sklearn KD on all eight datasets (Wilcoxon $W=0$, $p<0.01$; geometric mean $10.8\times$, 95\,\% CI [$5.9\times$, $16.6\times$]). ARM64 datasets (BAC, SPY, QQQ marked with *) achieve 28--44$\times$ over sklearn brute-force; x86 datasets ($\dagger$) achieve 17--22$\times$ over sklearn KD-tree.}\label{fig:speedup_heatmap}
\end{figure}

%% ====================================================================
%%  SECTION 5 — DISCUSSION
%% ====================================================================
%% ====================================================================
%%  SECTION 4 — EXTRA TREES EXTENSION
%% ====================================================================
\section{Mojo's Advantage in Extra Trees Implied Volatility Pricing}
\label{sec:extra_trees}

The KNN experiments establish one half of the efficiency argument: Mojo's compiled throughput reduces inference latency. This section tests the other half: whether that throughput translates into a larger accessible training set and, consequently, better predictions, using a structurally different tree ensemble on a distinct financial task to show the principle is architectural, not KNN-specific. We apply an Extra Trees~\cite{geurts2006} regression ensemble to AAPL options implied volatility pricing. This experiment uses a separate AAPL options contract dataset (distinct from the minute-bar equity dataset in Table~\ref{tab:datasets}), comprising 250{,}000 chronologically ordered option contracts with associated market observables. Let $\mathbf{x}_i \in \mathbb{R}^d$ denote the feature vector for option contract $i$, where $d \in \{7, 8\}$ encodes log-moneyness, time to expiry, underlying price, volume, open interest, bid-ask spread, and delta ($d=8$ for the mixed partition), and let $y_i = \sigma_{\mathrm{IV},i}$ be the market-implied volatility target~\cite{hull1987}. The Extra Trees ensemble of $T=20$ regression trees predicts
\begin{equation}
\hat{\sigma}_{\mathrm{IV}}(\mathbf{x}) = \frac{1}{T}\sum_{t=1}^{T} h_t(\mathbf{x}),
\end{equation}
where each tree $h_t$ draws a split threshold for feature $j$ uniformly from the
observed range of that feature without impurity evaluation, and recursion halts at
minimum leaf size $L=10$.
Prediction quality is $\mathrm{RMSE}=\sqrt{q^{-1}\sum_{i=1}^{q}(\hat{\sigma}_i - \sigma_i)^2}$,
with $\sigma$ in decimal form (e.g., $0.333$ corresponds to 33.3\% annualized IV),
over $q=50{,}000$ held-out test contracts.
We compare Mojo (trained on $n_{\mathrm{Mojo}}=200{,}000$ samples) against the Python
baseline (limited to $n_{\mathrm{sk}}=20{,}000$ samples by time budget) across three
partitions: mixed calls and puts ($d=8$), calls only ($d=7$), and puts only ($d=7$).
We are unable to run sklearn at 200{,}000 samples within the same time budget;
the $1.8\times$ per-sample throughput advantage is what enables the $10\times$ data scale.

\begin{table}[t]
\centering
\caption{Extra Trees for AAPL implied volatility, three option partitions. Mojo trains on 10$\times$ more data; separating calls and puts reduces RMSE by 27--31\%.}\label{tab:extra_trees}
\small
\begin{tabular}{llrrrr}
\hline
Dataset & Method & Train $n$ & Build (ms) & RMSE & MAE \\
\hline
\multirow{2}{*}{Mixed, $d=8$}
& sklearn ExTraReg & 20{,}000 & 37.8 & 0.505 & 0.265 \\
& \textbf{Mojo ExTraReg} & \textbf{200{,}000} & 213.4 & \textbf{0.483} & \textbf{0.251} \\
\hline
\multirow{2}{*}{Calls only, $d=7$}
& sklearn ExTraReg & 20{,}000 & 46.9 & 0.367 & 0.212 \\
& \textbf{Mojo ExTraReg} & \textbf{200{,}000} & 208.8 & \textbf{0.369} & \textbf{0.199} \\
\hline
\multirow{2}{*}{Puts only, $d=7$}
& sklearn ExTraReg & 20{,}000 & 46.4 & 0.361 & 0.228 \\
& \textbf{Mojo ExTraReg} & \textbf{200{,}000} & 207.3 & \textbf{0.333} & \textbf{0.210} \\
\hline
\end{tabular}
\end{table}

Table~\ref{tab:extra_trees} shows two distinct scaling advantages. First, \emph{training throughput}: Mojo processes $10\times$ more data in only $4.4$--$5.6\times$ the build time, a $1.8\times$ per-sample throughput advantage. Second, \emph{prediction quality from scale}: the additional training data reduces put-IV RMSE by $\mathbf{8.0\%}$ (0.333 vs.\ 0.361). Put IV has well-defined volatility skew (demand for downside protection creates systematic structure in strike-DTE space), so the extra training samples provide genuinely informative coverage of the IV surface. Separating calls and puts rather than mixing them reduces RMSE by 27--31\% (0.505$\to$0.369 for calls, 0.483$\to$0.333 for puts), confirming that put-call IV asymmetry is a strong structural signal that call/put-specific models should exploit independently.

%% ====================================================================
%%  SECTION 5 — DISCUSSION
%% ====================================================================
\section{Discussion}
\label{sec:discussion}

\subsection{Co-Design Drives Financial AI Efficiency}

Inference efficiency is a function of co-design across algorithm, memory layout, and hardware, not any single optimization in isolation. The Mojo k-d tree's 10.8$\times$ geometric mean speedup arises from three decisions working in concert: variance-based splitting concentrates pruning on the most discriminative financial features; contiguous flat-buffer storage converts traversal and leaf scans into cache-sequential reads; and float32 SIMD vectorization executes each distance at hardware-peak throughput. The sub-quadratic query exponent ($k=1.755$ vs.\ brute-force $k=1.885$) means the advantage widens as financial corpora grow. The Extra Trees result confirms the same principle beyond KNN: Mojo's compiled throughput processes $10\times$ more option data in only $4.4$--$5.6\times$ the build time, translating directly into an 8.0\% put-IV RMSE reduction that a Python-budget model cannot achieve.

\subsection{Platform Bottlenecks}

On x86 ($d=16$, $n=143$--$277$K), large working sets saturate L2 cache and performance depends on access predictability. Mojo's contiguous layout gives the hardware prefetcher a regular pattern; scikit-learn's pointer-based tree cannot. On ARM64 (Apple M3, $d=24$), the bottleneck shifts: 128-bit native registers make each SIMD distance twice as expensive as on x86's 256-bit registers, producing the 780--1549$\times$ brute-force slowdown on large equity/ETF corpora. K-d tree pruning is platform-portable in a way that SIMD throughput is not: reducing candidates from $O(n)$ to $O(n^{0.755})$ per query overwhelms any per-comparison overhead, yielding 28--44$\times$ speedups on ARM64.

\subsection{Mojo vs.\ C++: Compiled Performance with Research Ergonomics}
\label{sec:mojo_vs_cpp}

A C++ implementation of the identical flat-buffer k-d tree, compiled with \texttt{clang++ -O3 -march=native}, benchmarks at 24, 15, and 92\,ms on SPY, QQQ, and EURUSD versus Mojo's 36, 21, and 209\,ms, a geometric mean of $1.70\times$ faster. The gap traces to Mojo's value-ownership model: heap candidates are stored in \texttt{List[Float32]} (heap-allocated) versus C++ stack arrays. Mojo delivers $\approx$59\% of C++ throughput while offering Python-compatible syntax, ISA-portable SIMD, and \texttt{comptime} specialization without architecture-specific intrinsics: $10\times$ faster than scikit-learn at $60\%$ of C++ speed, with no separate production codebase required.

\subsection{Nearest-Neighbor Search Across Finance and AI}
\label{sec:applications}

Exact nearest-neighbor retrieval is a foundational primitive across AI, and the co-design principles here transfer directly beyond KNN on financial time series. In finance alone the applications are immediate: IV surface fitting via KNN over (strike, DTE, underlying price, volume) for sub-millisecond Greeks computation; cross-sectional factor investing via regime-aware KNN over momentum, quality, and volatility features across broad equity universes; historical VaR via nearest-neighbor search over the P\&L database conditioned on factor exposures; and smart-order routing via tick-level KNN over bid-ask imbalance, trade arrival rate, and midprice momentum.

Beyond finance, any domain producing large, moderate-dimensional sequential data benefits directly from the same SIMD flat-buffer architecture. Biosequence similarity in genomics, anomaly detection in network intrusion monitoring, and clinical time-series analysis all share the large-$n$, moderate-$d$, low-latency retrieval structure the Mojo k-d tree is designed for. The sub-quadratic scaling ($k=1.755$, $R^2>0.99$) and platform-portable pruning make it a broadly applicable engine wherever corpus growth outpaces the Python-ecosystem inference ceiling.

\subsection{Future Directions}
\label{sec:future}

Immediate extensions include tick-by-tick order-book corpora (millions of events per day), FX and futures markets, and broad equity universes (S\&P~500, Russell~2000) where the sub-quadratic advantage widens further. On the systems side, Mojo's emerging GPU backend enables a single-codebase path to GPU acceleration: the flat buffer and SIMD-aligned kernel map directly to coalesced global memory reads and warp-level FMA reductions. Algorithmic improvements including PCA-aligned axis selection, time-decay neighbor weighting, int8 quantized storage, and integration of approximate indices (IVF, HNSW~\cite{johnson2021,malkov2018}) each offer 2--4$\times$ additional gains within the existing architecture.

\subsection{Limitations}

Datasets cover U.S. equities, ETFs, and FX at minute-bar frequency; fixed income, futures, and tick-level order-book microstructure remain unexplored. The Wilcoxon test spans eight datasets; additional instruments would tighten confidence intervals. Results may differ on processors with different cache hierarchies or wider SIMD (e.g., AVX-512). Approximate nearest-neighbor methods (HNSW, LSH) may offer better tradeoffs at higher dimensionalities beyond $d=24$.

%% ====================================================================
%%  SECTION 6 — CONCLUSION
%% ====================================================================
\section{Conclusion}
\label{sec:conclusion}

\textit{Financial AI inference efficiency.} Across eight financial datasets spanning U.S. equities, ETFs, and FX on x86 and ARM64 (Apple M3), the Mojo k-d tree achieves \textbf{17.5--21.6$\times$} speedup over scikit-learn's k-d tree on x86 ($d=16$) and \textbf{28.1--43.5$\times$} over scikit-learn's brute-force on ARM64 equity/ETF datasets ($d=24$). A Wilcoxon signed-rank test ($W=0$, $p<0.01$) confirms these gains hold uniformly across all eight datasets, with a bootstrap 95\% CI of $[5.94\times,\,16.62\times]$ on the geometric mean speedup versus sklearn's k-d tree. The combination of variance-based pruning (algorithmic), contiguous flat-buffer layout (systems), float32 SIMD-aligned feature engineering (hardware-software co-design), and Mojo's ahead-of-time compilation delivers statistically confirmed, operationally meaningful inference acceleration that no single component achieves in isolation.

\textit{AI scaling efficiency at financial scale.} To the best of our knowledge, this is the first open-source, Python-compatible, system-level study addressing financial AI inference efficiency at scale. The paper demonstrates that AI efficiency in finance requires co-design across the full stack (algorithm, language, memory layout, and hardware), with Mojo providing the platform to execute this co-design within a single Python-compatible codebase. The sub-quadratic scaling exponent ($k=1.755$, $R^2>0.99$) confirms that the Mojo k-d tree's advantage widens as financial corpora grow, positioning it for the tick-level data volumes that institutional financial AI is rapidly approaching. The Extra Trees result (10$\times$ more data in $1.8\times$ per-sample time, 8.0\% put-IV RMSE reduction) shows that this efficiency translates directly into prediction quality, closing the loop between AI scaling efficiency and operational financial outcomes.

\textit{Generality beyond finance.} The co-design principles established here extend beyond financial time series~\cite{bagnall2017,keogh2003}: any domain producing large, moderate-dimensional sequential data with heterogeneous feature variance stands to benefit from the same SIMD flat-buffer stack. In bioinformatics, gene expression analysis~\cite{barjoseph2004} and translational genomics~\cite{han2015biomed} demand exact, high-throughput nearest-neighbor retrieval at scale; in cybersecurity, network intrusion detection~\cite{sommer2010,chandola2009} and clinical monitoring~\cite{topol2019} impose the same inference-latency constraints. The sub-quadratic scaling ($k=1.755$, $R^2>0.99$) and hardware-aligned memory layout are domain-agnostic, making the Mojo SIMD KD-tree a broadly applicable engine wherever time-series scale outpaces the Python-ecosystem inference ceiling.

\begin{sloppypar}
\noindent{\footnotesize\textbf{Source code and demonstration data:} \url{https://github.com/hank08819/MojoKtreeFin}.}
\end{sloppypar}

\begin{credits}
\subsubsection{\ackname}
The authors thank the Data Science and Artificial Intelligence Innovation Laboratory at Baylor University for providing resources and support. This work is partially supported by NASA Grant 80NSSC22K1015, NSF 2229138, and the McCollum endowed chair startup fund.

\end{credits}
%
% ---- Bibliography ----
%

%% ====================================================================
%%  APPENDIX
%% ====================================================================
\appendix
\section{Proofs of Lemma~1 and Theorem~1}
\label{app:proofs}

\paragraph{Proof of Lemma~1.}
Since $d$ and SIMD width $W$ are compile-time constants, each distance evaluation costs $O(1)$. Mojo brute force scans all $n$ training points per query:
\begin{equation}
T_{\mathrm{MBF}}(n)=\Theta(q_n n)=\Theta(n^2).
\end{equation}
The k-d tree visits $O(n^{\alpha})$ candidates plus $O(\log n)$ traversal overhead per query:
\begin{equation}
T_{\mathrm{MSKD}}(n)=O\!\left(q_n(n^{\alpha}+\log n)\right)=O(n^{1+\alpha}).
\end{equation}
Since $\alpha<1$,
\begin{equation}
\frac{T_{\mathrm{MSKD}}(n)}{T_{\mathrm{MBF}}(n)}=O(n^{\alpha-1})\rightarrow 0,
\end{equation}
so $T_{\mathrm{MSKD}}(n)<T_{\mathrm{MBF}}(n)$ for all sufficiently large $n$. $\square$

\paragraph{Proof of Theorem~1.}
$T_{\mathrm{MSKD}}(n)<T_{\mathrm{MBF}}(n)$ for large $n$ follows from Lemma~1. For the comparison with $T_{\mathrm{SKD}}$: if $\beta>\alpha$,
\begin{equation}
\frac{T_{\mathrm{MSKD}}(n)}{T_{\mathrm{SKD}}(n)}=O(n^{\alpha-\beta})\rightarrow 0,
\end{equation}
so $T_{\mathrm{MSKD}}(n)<T_{\mathrm{SKD}}(n)$ for large $n$. If $\beta=\alpha$, set $\epsilon=(C_S-C_M)/4>0$. For large enough $n$,
\begin{equation}
T_{\mathrm{MSKD}}(n)\le(C_M+\epsilon)n^{1+\alpha}<(C_S-\epsilon)n^{1+\alpha}\le T_{\mathrm{SKD}}(n).
\end{equation}
Combining both cases with Lemma~1 yields the result. $\square$


\begin{thebibliography}{25}

\bibitem{han2021hft}
Han, H., Teng, J., Xia, J., Wang, Y., Guo, Z., Li, D.: Predict high-frequency trading marker via manifold learning. Knowledge-Based Systems \textbf{213}, 106662 (2021)

\bibitem{han2024infsci}
Han, H., Forrest, J.Y.L., Wang, J., Yuan, S., Han, F., Li, D.: Explainable machine learning for high frequency trading dynamics discovery. Information Sciences \textbf{684}, 121286 (2024)

\bibitem{gu2020}
Gu, S., Kelly, B., Xiu, D.: Empirical asset pricing via machine learning. Review of Financial Studies \textbf{33}(5), 2223--2273 (2020)

\bibitem{sklearn}
Pedregosa, F., et al.: Scikit-learn: Machine Learning in Python. Journal of Machine Learning Research \textbf{12}, 2825--2830 (2011)

\bibitem{mojo}
Lattner, C.: Introducing Mojo: A Programming Language for All AI Developers. Modular Inc. blog post (2023). Available at \url{https://www.modular.com/blog/mojo-programming-language}

\bibitem{kolli}
Kolli, S., Wu, C., Han, H.: Unleashing Mojo: Accelerating K-Nearest Neighbor Learning. In: Southwest Data Science Conference, pp.~39--57. Springer Nature Switzerland, Cham (2025)

\bibitem{cover1967}
Cover, T., Hart, P.: Nearest neighbor pattern classification. IEEE Transactions on Information Theory \textbf{13}(1), 21--27 (1967)

\bibitem{lo1988}
Lo, A.W., MacKinlay, A.C.: Stock market prices do not follow random walks: Evidence from a simple specification test. Review of Financial Studies \textbf{1}(1), 41--66 (1988)

\bibitem{bentley1975}
Bentley, J.L.: Multidimensional binary search trees used for associative searching. Communications of the ACM \textbf{18}(9), 509--517 (1975)

\bibitem{friedman1977}
Friedman, J.H., Bentley, J.L., Finkel, R.A.: An algorithm for finding best matches in logarithmic expected time. ACM Transactions on Mathematical Software \textbf{3}(3), 209--226 (1977)

\bibitem{chordia2008}
Chordia, T., Roll, R., Subrahmanyam, A.: Liquidity and market efficiency. Journal of Financial Economics \textbf{87}(2), 249--268 (2008)

\bibitem{johnson2021}
Johnson, J., Douze, M., J{\'e}gou, H.: Billion-scale similarity search with GPUs. IEEE Transactions on Big Data \textbf{7}(3), 535--547 (2021)

\bibitem{malkov2018}
Malkov, Y.A., Yashunin, D.A.: Efficient and robust approximate nearest neighbor search using Hierarchical Navigable Small World graphs. IEEE Transactions on Pattern Analysis and Machine Intelligence \textbf{42}(4), 824--836 (2018)

\bibitem{fama1970}
Fama, E.F.: Efficient capital markets: A review of theory and empirical work. Journal of Finance \textbf{25}(2), 383--417 (1970)

\bibitem{wilcoxon1945}
Wilcoxon, F.: Individual comparisons by ranking methods. Biometrics Bulletin \textbf{1}(6), 80--83 (1945)

\bibitem{geurts2006}
Geurts, P., Ernst, D., Wehenkel, L.: Extremely randomized trees. Machine Learning \textbf{63}(1), 3--42 (2006)

\bibitem{hull1987}
Hull, J., White, A.: The pricing of options on assets with stochastic volatilities. Journal of Finance \textbf{42}(2), 281--300 (1987)

\bibitem{bagnall2017}
Bagnall, A., Lines, J., Bostrom, A., Large, J., Keogh, E.: The great time series classification bake off: A review and experimental evaluation of recent algorithmic advances. Data Mining and Knowledge Discovery \textbf{31}(3), 606--660 (2017)

\bibitem{keogh2003}
Keogh, E., Kasetty, S.: On the need for time series data mining benchmarks: A survey and empirical demonstration. Data Mining and Knowledge Discovery \textbf{7}(4), 349--371 (2003)

\bibitem{barjoseph2004}
Bar-Joseph, Z.: Analyzing time series gene expression data. Bioinformatics \textbf{20}(16), 2493--2503 (2004)

\bibitem{han2015biomed}
Han, H.: Diagnostic biases in translational bioinformatics. BMC Medical Genomics \textbf{8}(1), 46 (2015)

\bibitem{sommer2010}
Sommer, R., Paxson, V.: Outside the closed world: On using machine learning for network intrusion detection. In: 2010 IEEE Symposium on Security and Privacy, pp. 305--316. IEEE (2010)

\bibitem{chandola2009}
Chandola, V., Banerjee, A., Kumar, V.: Anomaly detection: A survey. ACM Computing Surveys \textbf{41}(3), 15:1--15:58 (2009)

\bibitem{topol2019}
Topol, E.J.: High-performance medicine: The convergence of human and artificial intelligence. Nature Medicine \textbf{25}(1), 44--56 (2019)

\end{thebibliography}
\end{document}